\documentclass{article}

\usepackage[preprint,nonatbib]{neurips_2024}

\usepackage[utf8]{inputenc} 
\usepackage[T1]{fontenc}    
\usepackage{hyperref}       
\usepackage{url}            
\usepackage{booktabs}       
\usepackage{amsfonts}       
\usepackage{nicefrac}       
\usepackage{microtype}      
\usepackage{xcolor}         

\usepackage{bm,multirow,graphicx,theorem,amssymb,amsmath,cleveref,subcaption,color,colortbl}
\usepackage[misc]{ifsym}

\title{Video-CCAM: Enhancing Video-Language Understanding with Causal Cross-Attention Masks for Short and Long Videos}

\author{
  Jiajun Fei$^1$\thanks{work done during internship at Tencent QQ, as a part of QQ MLLM project\\\indent$^\text{\Letter}$ corresponding author and project leader of QQ MLLM project}\ \ \ Dian Li$^{2,\ \text{\Letter}}$\ \ \ Zhidong Deng$^1$\ \ \ Zekun Wang$^2$\ \ \ Gang Liu $^2$\ \ \ Hui Wang$^2$\\ \\
  \text{Tsinghua University$^1$}\ \ \ \text{Tencent QQ$^2$}
}

\definecolor{Gray}{gray}{0.9}

\begin{document}

\maketitle

\begin{abstract}
  Multi-modal large language models (MLLMs) have demonstrated considerable potential across various downstream tasks that require cross-domain knowledge. MLLMs capable of processing videos, known as Video-MLLMs, have attracted broad interest in video-language understanding. However, videos, especially long videos, contain more visual tokens than images, making them difficult for LLMs to process. Existing works either downsample visual features or extend the LLM context size, risking the loss of high-resolution information or slowing down inference speed. To address these limitations, we apply cross-attention layers in the intermediate projector between the visual encoder and the large language model (LLM). As the naive cross-attention mechanism is insensitive to temporal order, we further introduce causal cross-attention masks (CCAMs) within the cross-attention layers. This Video-MLLM, named Video-CCAM, is trained in a straightforward two-stage fashion: feature alignment and visual instruction tuning. We develop several Video-CCAM models based on LLMs of different sizes (4B, 9B, and 14B). Video-CCAM proves to be a robust Video-MLLM and shows outstanding performance from short videos to long ones. Among standard video benchmarks like MVBench and VideoChatGPT-QA, Video-CCAM shows outstanding performances (1st/2nd/3rd in MVBench and TGIF-QA, 2nd/3rd/4th in MSVD-QA, MSRVTT-QA, and ActivityNet-QA). In benchmarks encompassing long videos, Video-CCAM models can be directly adapted to long video understanding and still achieve exceptional scores despite being trained solely with images and 16-frame videos. Using 96 frames (6$\times$ the training number of frames), Video-CCAM models rank 1st/2nd/3rd in VideoVista and 1st/2nd/4th in MLVU among all open-source Video-MLLMs, respectively. We provide a theoretical analysis of its temporal consistency and emphasize several key factors in its architecture through experiments. We hope that Video-CCAM can serve as a straightforward yet robust baseline for future Video-MLLM development. The code is publicly available in \url{https://github.com/QQ-MM/Video-CCAM}.
\end{abstract}

\section{Introduction}
\label{sec:intro}

Large language models (LLMs) such as GPT-4~\cite{openai2024gpt4}, Gemini~\cite{geminiteam2023gemini}, and LLaMA3~\cite{llama3}, have significantly reshaped the landscape of artificial intelligence, profoundly impacting our daily lives. These LLMs can engage in text-based conversations with users, meeting their needs and completing specific tasks~\cite{zhao2023survey}. Despite their potential as a step towards artificial general intelligence (AGI) assistants, their capabilities are confined to processing natural language. However, human interaction with the world is not limited to language alone; it also encompasses a variety of multi-modal information, such as vision, speech, audio, etc.

To address the language-only limitation, the research community has recently proposed various multi-modal large language models (MLLMs) that integrate additional modalities. Visual modality, especially image, has garnered considerable interest among all modalities. Notable developments include Flamingo~\cite{NEURIPS2022_960a172b}, which combines pre-trained vision and language models, exhibiting impressive multi-modal few-shot learning capabilities. MiniGPT-4~\cite{zhu2023minigpt4} aligns the visual encoder and Q-Former from BLIP-2~\cite{li2023blip2} with Vicuna~\cite{vicuna2023} through a single trainable projection layer, achieving advanced vision-language performance. LLaVA~\cite{liu2024visual} further introduces the concept of visual instruction tuning and showcases superior multi-modal abilities across various benchmarks. These pioneering approaches have collectively established a standard pipeline for MLLMs, typically including pre-trained large language models, modality-specific pre-trained encoders, trainable projectors, and datasets for feature alignment and instruction tuning. This framework has proven effective for integrating and leveraging image-text data.

The research field of MLLMs has recently seen a surge in Video-MLLMs~\cite{li2024videochat,maaz2023videochatgpt,luo2023valley,jin2023chatunivi,zhu2024languagebind,lin2023videollava,li2024mvbench}. Compared to images with two spatial dimensions, videos have an additional temporal dimension. Therefore, the number of visual tokens is not only related to the spatial resolution but also proportional to the number of video frames, which is difficult to accommodate within the limited context size of LLM. Existing works address this issue from modifying three components of Video-MLLMs, i.e., the LLM, the visual encoder, and the intermediate projector. Some works directly extend the context size of LLMs to hold more visual tokens. LWM~\cite{liu2024world} gradually increase the number of frames and the context size through multi-stage vision-language training. LongVA~\cite{zhang2024longva} first trains long-context LLM and then aligns it with images. However, the computational burden of long-context LLMs is significantly larger than normal ones. Other works adopt pooling~\cite{maaz2023videochatgpt}, downsampling~\cite{li2023llamavid,luo2023valley,zhang2024llavanextvideo}, or clustering~\cite{jin2023chatunivi} to directly reduce the number of visual tokens. These approaches are effective but come at the expense of fine-grained information loss. Unlike MLP projectors that do not alter the number of output tokens, cross-attention based projectors (Perceiver~\cite{NEURIPS2022_960a172b} and Q-Former~\cite{li2023blip2}) adopt a fixed number of queries to extract relevant information from visual inputs. For example, VideoChat2~\cite{li2024mvbench} uses 96 queries to process video inputs. However, cross-attention mechanism is insensitive to temporal order, which is crucial for accurate video understanding. Besides, these projectors generally have more parameters than MLP projectors.

In this work, we concentrate on the projector to address the abundant visual tokens. Our projector is centered around cross-attention layers, where a fixed number of queries is employed to process videos with different number of frames. This architecture makes it possible to handle extremely large number of video frames at no risk of exceeding the context length. Besides, we make several modifications to better process videos and simplify the training process. First, we propose causal cross-attention masks (CCAMs) within the cross-attention layer, making learnable queries temporally ordered and enhancing the model's video understanding ability. Second, we simplify the projector structure through reducing the number of layers and increasing the number of queries. We encapsulate our contributions as follows:

\begin{itemize}
  \item We propose Video-CCAM, an innovative Video-MLLM designed for advanced video-language understanding. Video-CCAM is a flexible model composed of a visual encoder, an LLM, and a projector, which employs cross-attention mechanism to process videos of variable frames and CCAMs to capture the temporal relationship within videos.
  \item We provide a theoretical analysis on the temporal consistency of CCAM. By treating videos as continuous signals, we demonstrate that the CCAM projector remains consistent for videos with different numbers of frames, making Video-CCAM a reliable Video-MLLM.
  \item We conduct extensive experiments to highlight Video-CCAM's outstanding performance. Among all open-source Video-MLLMs, Video-CCAM ranks 1st in MVBench~\cite{li2024mvbench}, 1st in VideoVista~\cite{li2024videovista}, 1st in MLVU~\cite{MLVU}, and 3rd in Video-MME~\cite{fu2024video}.
\end{itemize}

\section{Related Work}

\subsection{Image-MLLMs}

Images serve as a vital complement to the visual details that text alone cannot convey, thus playing a crucial role in multi-modal learning. Flamingo~\cite{NEURIPS2022_960a172b} leverages a pre-trained, frozen vision encoder to process input images and introduces the GATED XATTN-DENSE layer to integrate visual information into language models. However, the high training costs limit its accessibility. With the advancement of LLMs and the emergence of open-source projects~\cite{touvron2023llama,llama3,2023internlm,bai2023qwen}, there has been a proliferation of studies utilizing these powerful LLMs to create Image-MLLMs. BLIP-2~\cite{li2023blip2} introduces the Q-Former that connects frozen visual encoders with LLMs, facilitating various image-to-text tasks, including visual knowledge reasoning and conversation. MiniGPT-4~\cite{zhu2023minigpt4} further refines this approach by aligning the visual encoder and Q-Former in BLIP-2 with Vicuna~\cite{vicuna2023} through a single trainable projection layer, resulting in enhanced downstream performance. LLaVA~\cite{liu2024visual} expands the concept of instruction tuning to the visual domain, proposing visual instruction tuning as a follow-up to feature alignment pre-training. LLaVA and its successors~\cite{liu2023llavaplus,liu2023improvedllava,liu2024llavanext} demonstrate significant promise in addressing a variety of vision-language tasks. Recent works~\cite{ye2023mplugowl2,bai2023qwenvl,wang2024cogvlm} enhance the capabilities of MLLMs by innovating model architectures, introducing additional training stages, and curating high-quality training datasets, among other strategies.

\begin{figure*}[!t]
  \centering
  \includegraphics[width=\linewidth]{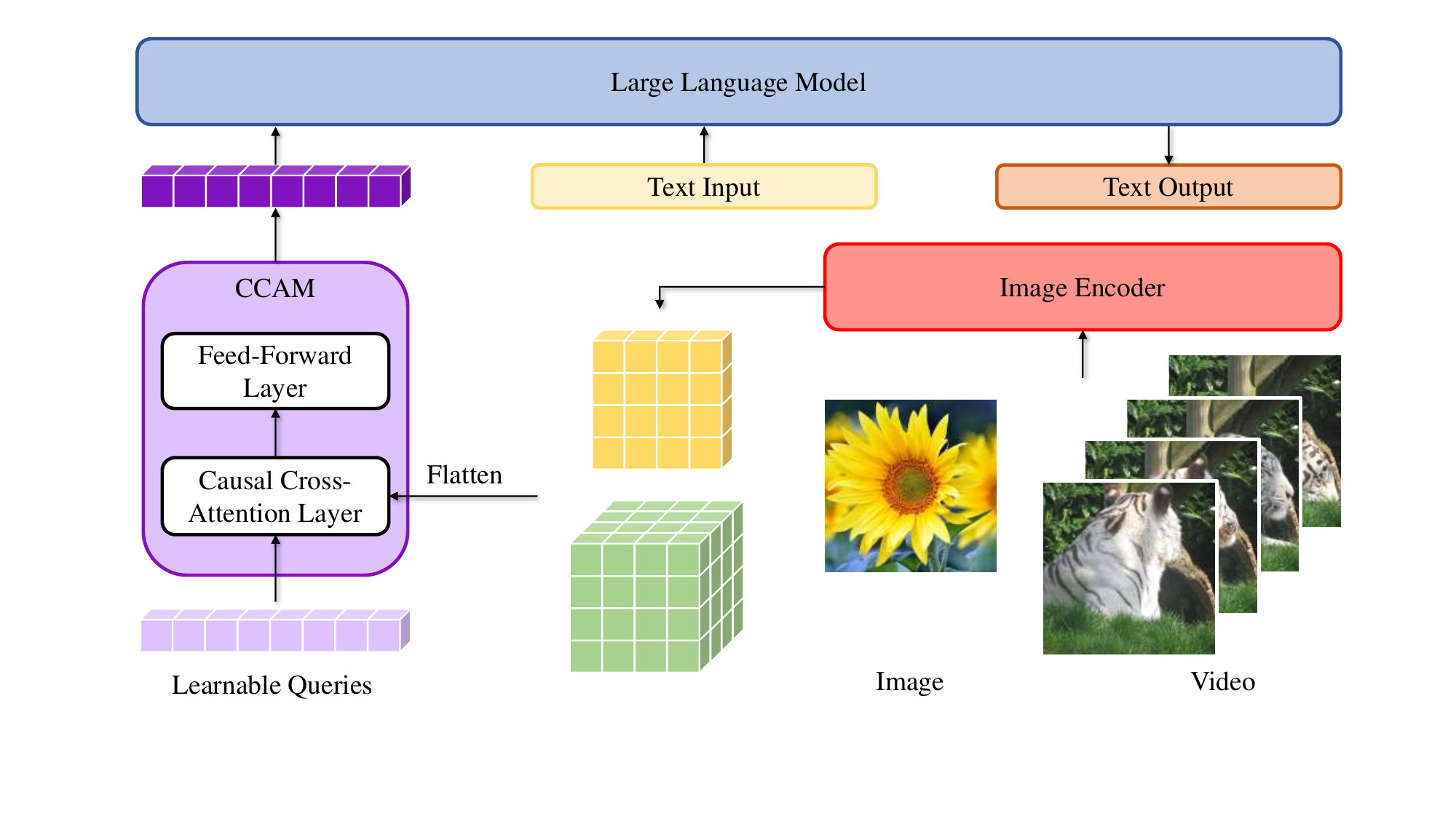}
  \caption{Overview of Video-CCAM. Video-CCAM adopts the same visual encoder to process images and video frames. Then, a collection of learnable queries distills the visual information. To preserve the temporal order of the video, CCAMs are implemented within the cross-attention layer, ensuring that Video-CCAM is aware of the chronological order of the video.}
  \label{fig:3}
\end{figure*}

\subsection{Video-MLLMs}
\label{sec:3}

As Image-MLLMs continue to mature, researchers are increasingly focusing on videos. Compared to images, videos have an additional temporal dimension, posing additional difficulties and challenges to Video-MLLMs. Similar to their image counterparts, Video-MLLMs primarily utilize two types of projectors: MLPs and Q-Formers~\cite{li2023blip2}. MLP projectors directly convert visual features from the encoder into embeddings. For instance, Video-ChatGPT~\cite{maaz2023videochatgpt} employs a linear layer to align spatially and temporally pooled video features with the LLM. PLLaVA~\cite{xu2024pllava} proposes an pooling strategy to reduce the domain differences between pre-trained image features and video ones. However, MLPs struggle to handle many frames, often forcing a trade-off between spatial resolution and temporal sampling density. Q-Formers output the same number of tokens as the number of learnable queries, independent of the input size. For example, VideoChat~\cite{li2024videochat} employs additional learnable queries to produce aligned visual embeddings. To address the Q-Former's limited frame differentiation, Vista-LLaMA~\cite{ma2023vistallama} recursively applies the Q-Former to model the temporal relationships. ST-LLM~\cite{liu2024stllm} also applies pre-trained Q-Formers on video frames to obtain compact visual representations. Beyond projectors, Video-MLLMs also face other challenges, particularly regarding the choice of video and image encoders. Since videos are often treated as sequences of images, most studies utilize image encoders to extract frame features, which are subsequently aggregated to represent video features. A majority of works, including VideoChat~\cite{li2024videochat}, Video-ChatGPT~\cite{maaz2023videochatgpt}, Valley~\cite{luo2023valley}, and Chat-UniVi~\cite{jin2023chatunivi}, employ CLIP ViT~\cite{pmlr-v139-radford21a} for processing both video frames and images. Others, such as LLaMA-VID~\cite{li2023llamavid}, TimeChat~\cite{ren2023timechat}, and Emu2~\cite{sun2023generative}, opt for EVA CLIP ViT~\cite{sun2023evaclip} as the visual encoder. Some researchers advocate that pre-trained video encoders are more suitable to capture temporal features. Video-LLaVA~\cite{lin2023videollava} underscores the significance of feature alignment across visual modalities and utilizes the visual encoders from LanguageBind~\cite{zhu2024languagebind} for processing visual inputs. UMT-L~\cite{Li_2023_ICCV}, a pre-trained video foundation model, is employed by VideoChat2~\cite{li2024mvbench} and has shown impressive performance across a range of downstream video-language tasks.

\section{Method}
\label{sec:4}

As illustrated in Fig.~\ref{fig:3}, Video-CCAM consists of three principal components: the visual encoder that processes images and videos, the LLM that handles visual and textual embeddings, and the CCAM projector that connects them.

\subsection{Visual Encoder}
\label{sec:1}

Existing Video-MLLMs generally employ three visual encoding strategies: using an image encoder, a video encoder, or both. In this work, we adopt image encoders for three reasons. Firstly, the generalization capabilities of pre-trained image encoders~\cite{pmlr-v139-radford21a,sun2023evaclip,Zhai_2023_ICCV} have been extensively validated, whereas the generalization capabilities of their video counterparts remain underexplored. Secondly, some video encoders have constraints on the number of input frames, whereas image encoders can be applied to arbitrary frames. Video-MLLMs built with these video encoders may give inaccurate responses if the input number of frames is different from that used during training. Lastly, Video-MLLMs with both image and video encoders require additional feature alignment efforts, which are not needed by those with a single encoder. Although most image encoders are not optimized for video processing, we argue that the autoregressive nature of LLMs can compensate for this limitation and enable them to interpret temporal visual tokens effectively.

\subsection{Projector}
\label{sec:2}

The projector is a crucial intermediary that connects the visual and textual embedding spaces in MLLMs. In this work, we focus on the projector, specifically the cross-attention based projector, to hold the large number of visual tokens in videos. However, naive cross-attention mechanism is insensitive to the temporal order within the video frames, since all queries can attend to all spatial and temporal visual tokens indiscriminately. For simplicity, we focus on one query embedding $Q_i\in\mathbb{R}^{1\times C} (0\leq i\leq N-1)$ and one attention head. We denote the key and value functions as $K, V:\mathbb{R}^{L\times C'}\rightarrow\mathbb{R}^{L\times C}$, respectively. For image embeddings with length as $L=H\times W$, the output of the cross-attention layer is computed as follows:
\begin{align}
  y_i=\frac{\exp\left(Q_iK^T\left(x\right)\right)V\left(x\right)}{\exp\left(Q_iK^T\left(x\right)\right)\bm{1}_{L}}\in\mathbb{R}^{1\times C},\label{equ:1}
\end{align}
where $x\in\mathbb{R}^{L\times C'}$ represents the image embeddings, and $\bm{1}_{L}=\left[1,\cdots,1\right]^T\in\mathbb{R}^{L\times 1}$ is a vector of ones. Subsequently, each query can integrate visual features from all positions. However, when it comes to video embeddings, each query considers visual features from not only all positions but also all moments:
\begin{align}
  y_i=\frac{\sum_j\exp\left(Q_iK^T\left(x_j\right)\right)V\left(x_j\right)}{\sum_j\exp\left(Q_iK^T\left(x_j\right)\right)\bm{1}_{L}},\label{equ:2}
\end{align}
where $\left[x_0, x_1, \cdots\right],x_i\in\mathbb{R}^{L\times C'}$ represents the video embeddings. Under these circumstances, it is possible that the initial queries may focus on the later visual embeddings while the last queries may concentrate on the earlier visual embeddings, which contradicts the LLM's autoregressive nature and may lead to incorrect responses with respect to the video inputs.

\begin{figure}[htbp]
  \centering
  \begin{minipage}{0.49\linewidth}
    \centering
    \includegraphics[width=\linewidth]{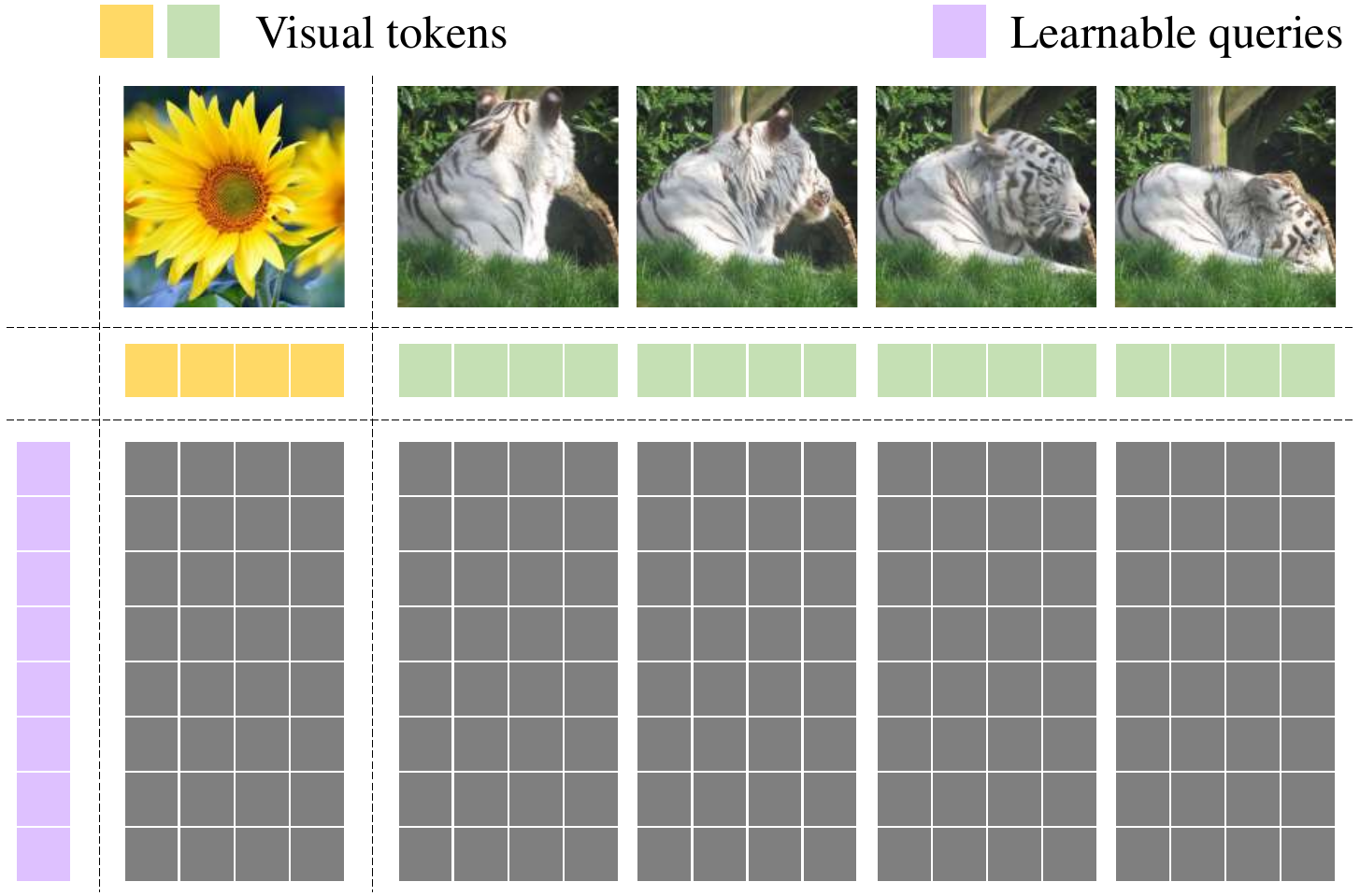}
    \subcaption{Conventional masks.}
    \label{fig:1}
  \end{minipage}
  \begin{minipage}{0.49\linewidth}
    \centering
    \includegraphics[width=\linewidth]{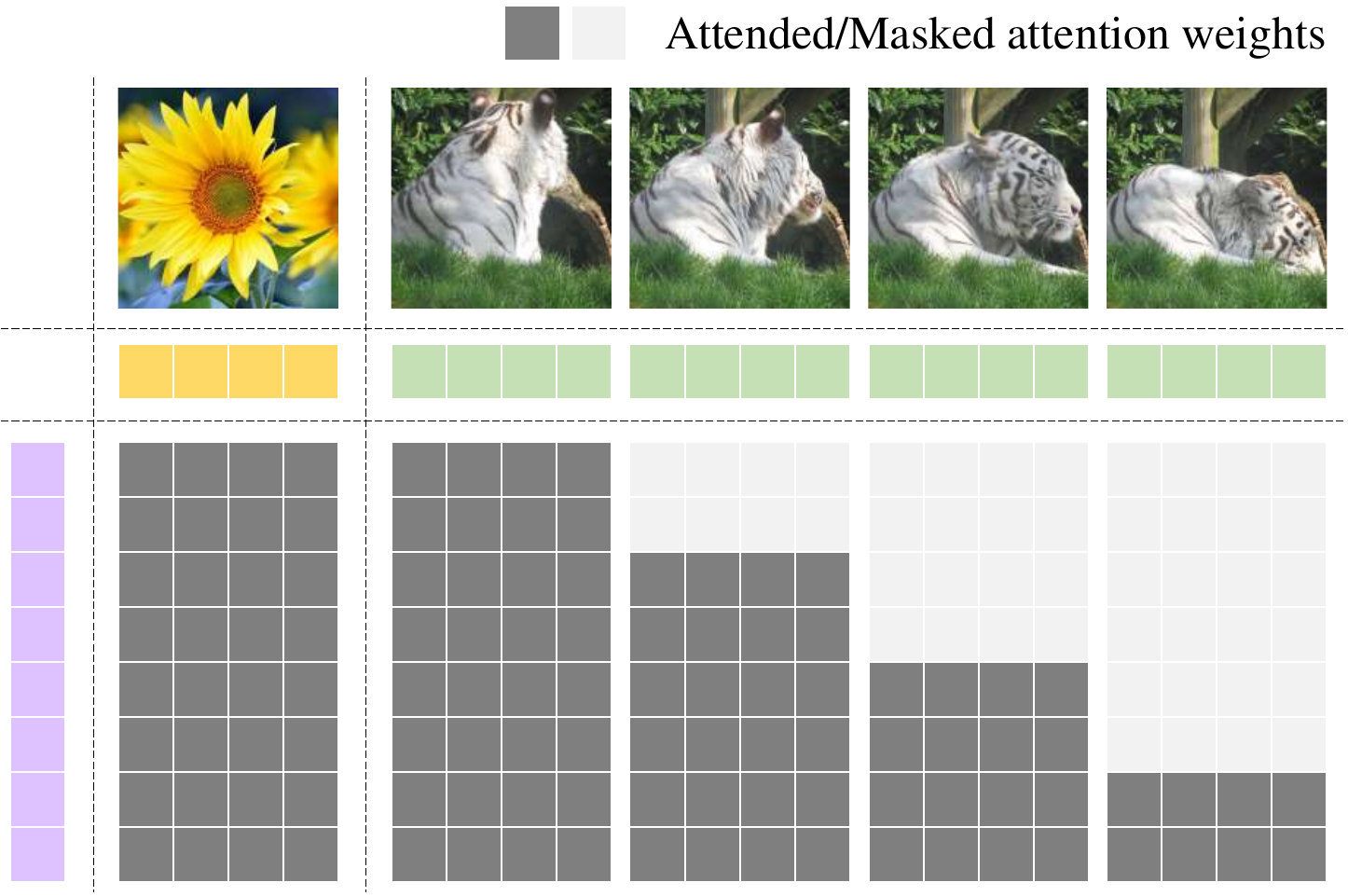}
    \subcaption{CCAM.}
    \label{fig:2}
  \end{minipage}
  \caption{Conventional cross-attention masks and CCAM. CCAM incrementally exposes video frames to learnable queries to decouple spatial and temporal features.}
  \label{fig:4}
\end{figure}

To mitigate this issue, we propose a simple approach by applying causal cross-attention masks (CCAMs), where the cross-attention output for video embeddings is computed as follows:
\begin{align}
  y_i=\frac{\sum_jM_{ij}\exp\left(Q_iK^T\left(x_j\right)\right)V\left(x_j\right)}{\sum_jM_{ij}\exp\left(Q_iK^T\left(x_j\right)\right)\bm{1}_{L}},\label{equ:5}
\end{align}
where $M_{ij}=1$ if the $i$-th query $Q_i$ is accessible to the $j$-th frame $x_j$. As $T$ is generally smaller than $N$, $M_{ij}=1$ if $i\geq j\left\lfloor\frac{N}{T}\right\rfloor$ else $M_{ij}=0$, and $\lfloor\cdot\rfloor$ is the floor function. We visualize the conventional cross-attention mask and our CCAM in Fig.~\ref{fig:4}. As depicted in Fig.~\ref{fig:1}, conventional cross-attention masks allow queries to attend to all visual tokens indiscriminately, which hinders the model's ability to discern temporal order across frames. In contrast, our CCAM, as illustrated in Fig.~\ref{fig:2}, ensures the initial queries focus on the early visual embeddings while allowing the last queries to access visual embeddings across different moments.

\subsection{Temporal consistency}
\label{sec:6}

In this work, we use temporal consistency to refer to the ability of the Video-MLLM to 1) effectively process videos with varying, often significantly larger, numbers of frames compared to those encountered during training; and 2) produce consistent outputs for the same video regardless of the number of sampled frames. Owing to the model structure and the training data distribution, some Video-MLLMs may encounter severe performance drops with different sampling strategies. However, a large number of frames is essential for long video understanding, which is significantly different from the training number of frames ($\leq 32$ for most Video-MLLMs). Therefore, existing works mostly employ an additional training stage to bridge this gap. Unlike these Video-MLLMs, Video-CCAM, despite being trained with images and 16-frame videos, can directly handle a large number of frames (e.g., 96 in VideoVista~\cite{li2024videovista}, MLVU~\cite{MLVU}, and Video-MME~\cite{fu2024video}) and shows outstanding performance without additional tuning. We attribute such experimental results to the temporal consistency of the CCAM projector, as illustrated below from the perspective of continuous signals.


First, we treat the video as a continuous signal instead of sampled frames, and then apply the visual encoder on the signal to get the visual embedding as $x\left(t\right):\left[0,T\right]\rightarrow\mathbb{R}^{L\times C'}$. Next, we replace the summation with the integral in \cref{equ:2} and gradually increase the upper limit of the integral to make the output sensitive to the temporal order:
\begin{align}
  y_i=\frac{\int_0^{T_i}\exp\left(Q_iK^T\left(x\left(\tau\right)\right)\right)V\left(x\left(\tau\right)\right)\text{d}\tau}{\int_0^{T_i}\exp\left(Q_iK^T\left(x\left(\tau\right)\right)\right)\bm{1}_{L}\text{d}\tau},\label{equ:7}
\end{align}
where $T_i=\frac{i+1}{N}T, 0\leq i\leq N-1$ and $T$ is the duration. Suppose that we sample one frame every $\Delta\tau$, then the discrete version of \cref{equ:7} becomes:
\begin{align}
  \tilde{y}_i=\frac{\sum_{j\Delta\tau\leq T_i}\exp\left(Q_iK^T\left(x_j\right)\right)V\left(x_j\right)\Delta\tau}{\sum_{j\Delta\tau\leq T_i}\exp\left(Q_iK^T\left(x_j\right)\right)\bm{1}_{L}\Delta\tau},\label{equ:8}
\end{align}
which is equivalent to \cref{equ:5} for $M_{ij}=\textbf{1}\left(j\Delta\tau\leq T_i\right)$. It is straightforward to prove that $\lim_{\Delta\tau\rightarrow 0}\tilde{y}_i=y_i$ if $K\left(\cdot\right),V\left(\cdot\right)$ are bounded (they are linear modules with bounded inputs in the implementation, so they are bounded). Given two differently sampled visual embeddings of the same video, their corresponding CCAM outputs are just approximations of \cref{equ:7} with different precision. In a word, the CCAM projector is able to not only handle videos of different length but also give consistent outputs for the same video with different numbers of sampled frames.

\subsection{Training Pipeline}

Video-CCAM is trained using a standard autoregressive loss, where the objective is to maximize the likelihood of the target textual outputs given the visual inputs and textual inputs. We take a simple two-stage training strategy. In the first pre-training stage, We randomly initialize the CCAM projector and leverage it to bridge the pre-trained visual encoder and LLM, both of which remain frozen. Only image-text data is utilized in this stage. In the second visual instruction tuning stage, more parameters in the visual encoder and LLM become tunable in addition to the projector. The instruction tuning dataset is composed of image-text and video-text pairs, thereby providing the model with richer context and more challenging tasks.


\section{Experiments}
\label{sec:5}

\subsection{Setup}
\label{sec:7}

\indent\indent\textbf{Model.} We use SigLIP-SO400M~\cite{Zhai_2023_ICCV} as the visual encoder and conduct experiments on three LLMs, i.e., Phi-3-mini-4k-instruct~\cite{abdin2024phi3} (4B), Yi-1.5-9B-Chat~\cite{ai2024yiopenfoundationmodels}, and Phi-3-medium-4k-instruct~\cite{abdin2024phi3} (14B). The resulting models are denoted as Video-CCAM-4B, Video-CCAM-9B, and Video-CCAM-14B. Our implementation is based on the xtuner~\cite{2023xtuner} repository. In the first stage, only the projector is tuned. In the second stage, we incorporate LoRA~\cite{hu2022lora} on the visual encoder and the LLM. All CCAMs are composed of one causal cross-attention layer and one feed-forward layer with 1,024 learnable queries.

\textbf{Dataset.} In the first stage, we use the LCS-558K~\cite{liu2024visual} for alignment. In the second stage, we combine the instruction tuning datasets of VideoChat2~\cite{li2024mvbench} and LLaVA-Hound~\cite{zhang2024directpreferenceoptimizationvideo}. To enrich the data diversity, we further add several question answering and caption datasets (the training split), including EgoTaskQA~\cite{jia2022egotaskqa}, PerceptionTestQA~\cite{patraucean2023perception}, ActivityNetQA~\cite{10.1609/aaai.v33i01.33019127}, STAR~\cite{wu2021star}, etc. For short or incomplete responses, some are abandoned while the others are rephrased into long and complete sentences by GPT-4o-mini~\cite{gpt4omini} and Gemini 1.5 Flash~\cite{geminiteam2024gemini15unlockingmultimodal}. Finally, we get 4.4M samples in total. Video-CCAM is trained for 1 epoch with images and 16-frame videos. All experiments are done with $8\times$ NVIDIA H800 GPUs. The total training duration of Video-CCAM-4B and Video-CCAM-14B are 2.5 days and 6 days, respectively.

\textbf{Evaluation.} We evaluate our Video-CCAMs with several benchmarks, i.e., MVBench~\cite{li2024mvbench}, VideoVista~\cite{li2024videovista}, MLVU~\cite{MLVU}, VideoChatGPT-QA~\cite{maaz2023videochatgpt}, and Video-MME~\cite{fu2024video}. As shown in \Cref{tab:4}, the videos in all benchmarks except MVBench~\cite{li2024mvbench} are significantly longer than those in the training data on average, posing great challenges on our Video-CCAM models.

\begin{table}[htbp]
  \centering
  \caption{Video duration in the training data and benchmarks. All values are in seconds.}
  \label{tab:4}
  \begin{tabular}{cccccc}
    \hline
    Name & Mean & Min & Max & Median & 95 Percentile \\
    \hline
    Train & 22.8 & 0.4 & 755.0 & 10.9 & 86.2 \\
    \hline
    MVBench~\cite{li2024mvbench} & 26.7 & 1.0 & 527.0 & 14.2 & 117.2 \\
    VideoVista~\cite{li2024videovista} & 152.1 & 0.8 & 918.5 & 96.9 & 594.1 \\
    MLVU~\cite{MLVU} & 704.6 & 180.0 & 32550.1 & 480.0 & 1222.8 \\
    Video-MME~\cite{fu2024video} & 1020.5 & 11.0 & 3579.4  & 487.9 & 3039.0 \\
    \hline
  \end{tabular}
\end{table}

\begin{table*}[!t]
  \centering
  \caption{Evaluation results in MVBench~\cite{li2024mvbench}. All Image-MLLMs concat 4 frame embeddings before feeding into the LLM~\cite{dai2023instructblip}. All Video-MLLMs are evaluated with 16 frames except VideoChatGPT~\cite{maaz2023videochatgpt} (100 frames), Video-CCAM (32 frames). The best and second best results are \textbf{bold} and \underline{underlined}, respectively. Sub-task names are abbreviated to improve readability.}
  \label{tab:1}
  \addtolength{\tabcolsep}{-2.2pt}
  \begin{tabular}{ccccccccccccc}
    \hline
    \multirow{2}{*}{Model} & LLM & \multirow{2}{*}{Mean} & AA & AC & AL & AP & AS & CO & CI & EN & ER & FA\\
    & Size & & FP & MA & MC & MD & OE & OI & OS & ST & SC & UA \\
    \hline
    \multirow{2}{*}{Random} & \multirow{2}{*}{-} &\multirow{2}{*}{28.00}&33.3 & 33.3 & 25.0 & 25.0 & 25.0 & 33.3 & 30.9 & 25.0 & 20.0 & 25.0 \\
    & & & 25.0 & 33.3 & 25.0 & 25.0 & 33.3 & 25.0 & 33.3 & 25.0 & 33.3 & 25.0 \\
    LLaMA-Adapter & \multirow{2}{*}{7B} & \multirow{2}{*}{31.70} & 51.0 & 29.0 & 21.5 & 28.0 & 23.0 & 31.5 & 32.0 & 22.5 & 28.0 & 30.0 \\
    \cite{zhang2024llamaadapter} & & & 25.0 & 41.5 & 22.5 & 25.5 & 53.5 & 32.5 & 33.5 & 30.5 & 39.5 & 33.0 \\
    VideoChatGPT & \multirow{2}{*}{7B} & \multirow{2}{*}{32.70} & 62.0 & 30.5 & 20.0 & 26.0 & 23.5 & 33.0 & 35.5 & 29.5 & 26.0 & 22.5 \\
    \cite{maaz2023videochatgpt} & & & 29.0 & 39.5 & 25.5 & 23.0 & 54.0 & 28.0 & 40.0 & 31.0 & 48.5 & 26.5 \\
    Video-LLaMA & \multirow{2}{*}{7B} & \multirow{2}{*}{34.10} & 51.0 & 34.0 & 22.5 & 25.5 & 27.5 & 40.0 & 37.0 & 30.0 & 21.0 & 29.0 \\
    \cite{zhang-etal-2023-video} & & & 32.5 & 32.5 & 22.5 & 22.5 & 48.0 & 40.5 & 38.0 & 43.0 & 45.5 & 39.0 \\
    VideoChat & \multirow{2}{*}{7B} & \multirow{2}{*}{34.10} & 56.0 & 35.0 & 27.0 & 26.5 & 33.5 & 41.0 & 36.0 & 23.5 & 23.5 & 33.5 \\
    \cite{li2024videochat} & & & 26.5 & 42.5 & 20.5 & 25.5 & 53.0 & 40.5 & 30.0 & 48.5 & 46.0 & 40.5 \\
    LLaVA & \multirow{2}{*}{7B} & \multirow{2}{*}{36.00} & 63.0 & 34.0 & 20.5 & 39.5 & 28.0 & 36.0 & 42.0 & 27.0 & 26.5 & 30.5 \\
    \cite{liu2024visual} & & & 25.0 & 38.5 & 20.5 & 23.0 & 53.0 & 41.0 & 41.5 & 45.0 & 47.0 & 39.0 \\
    VideoChat2 & \multirow{2}{*}{7B} & \multirow{2}{*}{51.10} & 83.5 & 39.0 & 23.0 & 47.5 & 66.0 & 36.5 & 65.5 & 35.0 & 40.5 & 49.5 \\
    \cite{li2024mvbench} & & & 49.0 & 58.5 & 42.0 & 23.0 & 58.0 & 71.5 & 42.5 & 88.5 & 44.0 & 60.0 \\
    ST-LLM & \multirow{2}{*}{7B} & \multirow{2}{*}{54.85} & 84.0 & 36.0 & 31.0 & 53.5 & 66.0 & 45.5 & 58.0 & 34.5 & 41.5 & 44.0 \\
    \cite{liu2024stllm} & & & 44.5 & 78.0 & 57.0 & 43.0 & 80.5 & 73.5 & 39.0 & 86.5 & 42.5 & 58.5 \\
    PLLaVA 34B & & & 82.0 & 40.5 & 49.5 & 53.0 & 67.5 & 66.5 & 59.0 & 39.5 & 63.5 & 47.0 \\
    \cite{xu2024pllava} & \multirow{-2}{*}{34B} & \multirow{-2}{*}{58.13} & 50.0 & 70.0 & 43.0 & 37.5 & 68.5 & 67.5 & 36.5 & 91.0 & 51.5 & 79.0 \\
    VideoChat2 HD & & & 79.5 & 60.0 & 87.5 & 50.0 & 68.5 & 93.5 & 71.5 & 36.5 & 45.0 & 49.5 \\
    \cite{li2024mvbench} & \multirow{-2}{*}{7B} & \multirow{-2}{*}{62.30} & 87.0 & 40.0 & 76.0 & 92.0 & 53.0 & 62.0 & 45.5 & 36.0 & 44.0 & 69.5 \\
    \hline
    \multirow{2}{*}{Video-CCAM-4B} & \multirow{2}{*}{4B} & \multirow{2}{*}{62.80} & 85.5 & 56.5 & 32.5 & 61.0 & 81.5 & 75.0 & 58.0 & 30.5 & 67.0 & 52.0 \\
    & & & 51.5 & 79.5 & 57.5 & 26.0 & 79.5 & 81.5 & 47.0 & 90.5 & 65.0 & 78.5 \\
    \multirow{2}{*}{Video-CCAM-9B} & \multirow{2}{*}{9B} & \multirow{2}{*}{\textbf{64.60}} & 89.5 & 59.0 & 29.0 & 67.0 & 83.0 & 77.0 & 59.0 & 34.0 & 73.5 & 49.0 \\
    & & & 54.0 & 85.0 & 67.0 & 28.0 & 86.5 & 81.0 & 45.0 & 90.0 & 63.5 & 72.0  \\
    \multirow{2}{*}{Video-CCAM-14B} & \multirow{2}{*}{14B} & \multirow{2}{*}{\underline{63.08}} & 88.0 & 59.0 & 38.5 & 66.0 & 84.5 & 76.5 & 52.5 & 29.0 & 79.0 & 47.0 \\
    & & & 54.0 & 74.5 & 57.0 & 21.5 & 71.0 & 85.0 & 40.0 & 90.5 & 68.5 & 79.5 \\
    \hline
  \end{tabular}
  \addtolength{\tabcolsep}{2.2pt}
\end{table*}

\begin{table}[!b]
  \centering
  \caption{Evaluation results in VideoVista~\cite{li2024videovista}. The best and second best results among open-source MLLMs are \textbf{bold} and \underline{underlined}, respectively.}
  \label{tab:10}
  \addtolength{\tabcolsep}{-1pt}
  \begin{tabular}{cccccc}
    \hline
    Model & LLM Size & Frames & Overall & Understanding & Reasoning \\
    \hline
    \multicolumn{6}{c}{\textit{Open-source MLLMs}} \\
    \hline
    VideoChatGPT~\cite{maaz2023videochatgpt} & 7B & 100 & 36.65 & 36.09 & 38.73 \\
    Video-LLaVA~\cite{lin2023videollava} & 7B & 8 & 56.59 & 53.82 & 66.91 \\
    LLaVA-NeXT-Video~\cite{zhang2024llavanextvideo} & 7B & 16 & 56.66 & 54.12 & 66.14 \\
    LLaMA-VID~\cite{li2023llamavid} & 7B & 1 FPS & 56.87 & 54.00 & 67.61 \\
    VideoChat2 HD~\cite{li2024mvbench} & 7B & 16 & 61.58 & 59.27 & 70.24 \\
    VILA-1.5~\cite{lin2023vila} & 13B & 8 & 64.18 & 62.27 & 71.34 \\
    LongVA~\cite{zhang2024longva} & 7B & 128 & 67.36 & 64.67 & 77.39 \\
    InternLM-XComposer-2.5~\cite{zhang2024internlmxcomposer25versatilelargevision} & 7B & 64 & 68.91 & 66.75 & 76.96 \\
    \hline
    \multicolumn{6}{c}{\textit{Close-source MLLMs}} \\
    \hline
    GPT-4o-mini~\cite{gpt4omini} & - & 100 & 75.76 & 72.87 & 85.52 \\
    Gemini 1.5 Flash~\cite{geminiteam2024gemini15unlockingmultimodal} & - & 1 FPS & 76.39 & 74.73 & 82.30 \\
    GPT-4o~\cite{gpt4o} & - & 100 & 78.26 & 75.15 & 87.97 \\
    \hline
    \multicolumn{6}{c}{\textit{Video-CCAM}} \\
    \hline
    Video-CCAM-4B & 4B & 96 & \underline{70.82} & 67.49 & 82.31 \\
    Video-CCAM-9B & 9B & 96 & 69.00 & 65.55 & 80.92 \\
    Video-CCAM-14B & 14B & 96 & \textbf{76.55} & 73.54 & 86.99 \\
    \hline
  \end{tabular}
  \addtolength{\tabcolsep}{1pt}
\end{table}

\subsection{MVBench~\cite{li2024mvbench}}

MVBench~\cite{li2024mvbench} is a comprehensive benchmark that includes 20 distinct video tasks, each with 200 questions designed to probe the model's understanding of video content. As shown in \Cref{tab:1}, Video-CCAM-4B surpasses all previous MLLMs despite its small size, demonstrating its efficiency and effectiveness in handling video-language understanding tasks. Meanwhile, Video-CCAM-9B sets a new SOTA result, further showcasing its superior performance in this benchmark.

\subsection{VideoVista~\cite{li2024videovista}}

VideoVista~\cite{li2024videovista} is another comprehensive benchmark tailored for video understanding and reasoning, including 3,400 videos and 25,000 questions across 14 categories. As the experimental results in \Cref{tab:10} show, Video-CCAM-4B surpasses all previous open-source Video-MLLMs, while Video-CCAM-14B sets a new SOTA result among open-source Video-MLLMs and demonstrates similar performance to GPT-4o-mini~\cite{gpt4omini} and Gemini 1.5 Flash~\cite{geminiteam2024gemini15unlockingmultimodal}.

\subsection{MLVU~\cite{MLVU}}

MLVU~\cite{MLVU} is a long video understanding benchmark with 9 distinct tasks divided into Multi-Choice (M) and Generation (G) categories. For the Generation tasks, MLVU utilizes GPT-4-Turbo~\cite{gpt4turbo} to assign scores to model responses. While Video-CCAM models do not achieve top results, the performance gaps between them and the best open-source results are small. For the Multi-Choice tasks, Video-CCAM-4B is comparable to previous open-source SOTA Video-MLLMs, while Video-CCAM-14B sets a new SOTA result among open-source Video-MLLMs. Despite the duration differences between training data and MLVU~\cite{MLVU} in \Cref{tab:4}, Video-CCAM is still proficient at handling long video understanding.

\begin{table}[htbp]
  \centering
  \caption{Evaluation results in MLVU~\cite{MLVU}. The best and second best results are \textbf{bold} and \underline{underlined}, respectively.}
  \label{tab:9}
  \begin{tabular}{cccc}
    \hline
    Model & Frames & M-Avg & G-Avg \\
    \hline
    VideoChatGPT~\cite{maaz2023videochatgpt} & 100 & 31.3 & 3.90 \\
    LLaMA-VID~\cite{li2023llamavid} & 1 FPS & 33.2 & 4.22 \\
    LLaVA-NeXT-Video~\cite{zhang2024llavanextvideo} & 16 & 39.3 & 3.23 \\
    Qwen-VL-Max~\cite{bai2023qwenvl} & 16 & 42.2 & 3.96 \\
    Video-LLaVA~\cite{lin2023videollava} & 8 & 47.3 & 3.84 \\
    VideoChat2 HD~\cite{li2024mvbench} & 16 & 47.9 & 3.99 \\
    LongVA~\cite{zhang2024longva} & 256 & 56.3 & \underline{4.33} \\
    VILA-1.5~\cite{lin2023vila} & 14 & 56.7 & 4.31 \\
    GPT-4o~\cite{gpt4o} & 0.5 FPS & \textbf{64.6} & \textbf{5.80} \\
    \hline
    Video-CCAM-4B & 96 & 56.5 & 4.09 \\
    Video-CCAM-9B & 86 & 58.5 & 3.98 \\
    Video-CCAM-14B & 96 & \underline{63.1} & 4.01 \\
    \hline
  \end{tabular}
\end{table}

\subsection{VideoChatGPT-QA~\cite{maaz2023videochatgpt}}

\begin{table*}[!t]
  \centering
  \caption{Evaluation results in VideoChatGPT-QA~\cite{maaz2023videochatgpt}. The best and second best results among open-source MLLMs are \textbf{bold} and \underline{underlined}, respectively. Video-CCAM models are evaluated with 32 frames.}
  \label{tab:2}
  \addtolength{\tabcolsep}{-2pt}
  \begin{tabular}{cccccccccc}
    \hline
    \multirow{2}{*}{Method} & LLM & \multicolumn{2}{c}{MSVD-QA} & \multicolumn{2}{c}{MSRVTT-QA} & \multicolumn{2}{c}{ActivityNet-QA} & \multicolumn{2}{c}{TGIF-QA} \\
    & Size & Acc. & Score & Acc. & Score & Acc. & Score & Acc. & Score \\
    \hline
    LLaMA-Adapter~\cite{zhang2024llamaadapter} & 7B & 54.9 & 3.1 & 43.8 & 2.7 & 34.2 & 2.7 & - & - \\
    Video-LLaMA~\cite{zhang-etal-2023-video} & 7B & 51.6 & 2.5 & 29.6 & 1.8 & 12.4 & 1.1 & - & - \\
    VideoChatGPT~\cite{maaz2023videochatgpt} & 7B & 64.9 & 3.3 & 49.3 & 2.8 & 35.2 & 2.7 & 51.4 & 3.0 \\
    Video-LLaVA~\cite{lin2023videollava} & 7B & 70.7 & 3.9 & 59.2 & 3.5 & 45.3 & 3.3 & 70.0 & 4.0 \\
    Chat-UniVi~\cite{jin2023chatunivi} & 7B & 65.0 & 3.6 & 54.6 & 3.1 & 45.8 & 3.2 & 60.3 & 3.4 \\
    VideoChat~\cite{li2024videochat} & 7B & 56.3 & 2.8 & 45.0 & 2.5 & 26.5 & 2.2 & 34.4 & 2.3 \\
    VideoChat2~\cite{li2024mvbench} & 7B & 70.0 & 3.9 & 54.1 & 3.3 & 49.1 & 3.3 & - & - \\
    Vista-LLaMA~\cite{ma2023vistallama} & 7B & 65.3 & 3.6 & 60.5 & 3.3 & 48.3 & 3.3 & - & - \\
    LLaMA-VID~\cite{li2023llamavid} & 13B & 70.0 & 3.7 & 58.9 & 3.3 & 47.5 & 3.3 & - & - \\
    ST-LLM~\cite{liu2024stllm} & 7B & 74.6 & 3.9 & 63.2 & 3.4 & 50.9 & 3.3 & - & - \\
    PLLaVA~\cite{xu2024pllava} & 34B & \textbf{79.9} & \textbf{4.2} & \textbf{68.7} & \textbf{3.8} & \textbf{60.9} & \underline{3.7} & 80.6 & 4.3 \\
    \hline
    Video-CCAM-4B & 4B & 76.9 & \underline{4.1} & 64.4 & \underline{3.7} & 58.0 & \underline{3.7} & 83.0 & \underline{4.4} \\
    Video-CCAM-9B & 9B & 77.9 & \textbf{4.2} & 65.9 & \textbf{3.8} & 59.7 & \textbf{3.8} & \underline{84.0} & \textbf{4.5} \\
    Video-CCAM-14B & 14B & \underline{78.6} & \textbf{4.2} & \underline{66.3} & \textbf{3.8} & \underline{60.4} & \textbf{3.8} & \textbf{84.4} & \textbf{4.5} \\
    \hline
  \end{tabular}
  \addtolength{\tabcolsep}{2pt}
\end{table*}

VideoChatGPT-QA~\cite{maaz2023videochatgpt} encompasses a variety of validation/test datasets from MSRVTT-QA~\cite{10.1145/3123266.3123427}, MSVD-QA~\cite{10.1145/3123266.3123427}, TGIF-QA~\cite{Jang_2017_CVPR}, and ActivityNet-QA~\cite{10.1609/aaai.v33i01.33019127}. Following VideoChatGPT~\cite{maaz2023videochatgpt}, we employ GPT-3.5-Turbo~\cite{gpt35turbo} to evaluate the predictions. As shown in \Cref{tab:2}, Video-CCAM-4B outperforms all previous works except PLLaVA-34B~\cite{xu2024pllava}, and Video-CCAM-14B further closes the gap between medium-sized Video-MLLMs and PLLaVA-34B~\cite{xu2024pllava}. Notably, both Video-CCAM models have better accuracies and scores in TGIF-QA~\cite{Jang_2017_CVPR} than all previous models.

\subsection{Video-MME~\cite{fu2024video}}

Video-MME~\cite{fu2024video} is another comprehensive multi-modal evaluation benchmark for Video-MLLMs, offering a highly diverse range of video types and temporal durations and posing significant challenges for Video-MLLMs trained with few frames. As shown in \Cref{tab:8}, Video-CCAM-4B demonstrates competitive performance and is only slightly weaker than the much larger InternVL-Chat-V1.5~\cite{chen2024far} and Qwen-VL-Max~\cite{bai2023qwenvl}. Video-CCAM-14B ranks the highest among all open-source MLLMs with fewer than 30B parameters.

\begin{table*}[htbp]
  \centering
  \caption{Evaluation results in Video-MME~\cite{fu2024video}. 'w/o s'/'w s' stands for 'without/with subtitles.}
  \label{tab:8}
  \addtolength{\tabcolsep}{-2.5pt}
  \begin{tabular}{ccccccccccc}
    \hline
    \multirow{2}{*}{Method} & LLM & \multirow{2}{*}{Frames} & \multicolumn{2}{c}{Overall (\%)} & \multicolumn{2}{c}{Short (\%)} & \multicolumn{2}{c}{Medium (\%)} & \multicolumn{2}{c}{Long (\%)} \\
    & Size & & w/o s & w s & w/o s & w s & w/o s & w s & w/o s & w s \\
    \hline
    \multicolumn{11}{c}{\textit{Open-source MLLMs}} \\
    \hline
    Video-LLaVA~\cite{lin2023videollava} & 7B & 8 & 39.9 & 41.6 & 45.3 & 46.1 & 38.0 & 40.7 & 36.2 & 38.1 \\
    ST-LLM~\cite{liu2024stllm} & 7B & 64 & 37.9 & 42.3 & 45.7 & 48.4 & 36.8 & 41.4 & 31.3 & 36.9 \\
    InternVL-Chat-V1.5~\cite{chen2024far} & 20B & 10 & 50.7 & 52.4 & 60.2 & 61.7 & 46.4 & 49.1 & 45.6 & 46.6 \\
    LongVA~\cite{zhang2024longva} & 7B & 128 & 52.6 & 54.3 & 61.1 & 61.6 & 50.4 & 53.6 & 46.2 & 47.6 \\
    VILA-1.5~\cite{lin2023vila} & 34B & 14 & 60.1 & 61.1 & 68.7 & 69.9 & 58.8 & 59.7 & 53.0 & 53.8 \\
    LLaVA-NeXT-Video~\cite{zhang2024llavanextvideo} & 32B & 32 & 60.2 & 63.0 & 73.2 & 76.0 & 57.0 & 59.7 & 50.3 & 53.3 \\
    \hline
    \multicolumn{11}{c}{\textit{Close-source MLLMs}} \\
    \hline
    Qwen-VL-Max~\cite{bai2023qwenvl} & - & 4 & 51.3 & 51.2 & 55.8 & 57.6 & 49.2 & 48.9 & 48.9 & 47.0 \\
    GPT-4V~\cite{gpt4v} & - & 10 & 59.9 & 63.3 & 70.5 & 73.2 & 55.8 & 59.7 & 53.5 & 56.9 \\
    Gemini 1.5 Flash~\cite{geminiteam2024gemini15unlockingmultimodal} & - & 1/2 FPS & 70.3 & 75.0 & 78.8 & 79.8 & 68.8 & 74.7 & 61.1 & 68.8 \\
    GPT-4o~\cite{gpt4o} & - & 384 & 71.9 & 77.2 & 80.0 & 82.8 & 70.3 & 76.6 & 65.3 & 72.1 \\
    Gemini 1.5 Pro~\cite{geminiteam2024gemini15unlockingmultimodal} & - & 1/2 FPS & 75.0 & 81.3 & 81.7	& 84.5 & 74.3 & 81.0 & 67.4 & 77.4 \\
    \hline
    \multicolumn{11}{c}{\textit{Video-CCAM}} \\
    \hline
    Video-CCAM-4B & 4B & 96 & 50.1 & 51.2 & 59.6 & 58.9 & 49.9 & 51.4 & 40.9 & 43.5 \\
    Video-CCAM-9B & 9B & 96 & 50.3 & 52.6 & 61.9 & 63.1 & 49.2 & 52.3 & 39.6 & 42.4 \\
    Video-CCAM-14B & 14B & 96 & 53.9 & 56.1 & 62.1 & 63.9 & 52.8 & 55.9 & 47.0 & 48.3 \\
    \hline
  \end{tabular}
  \addtolength{\tabcolsep}{2.5pt}
\end{table*}

\subsection{Ablation Study}

We conduct several ablation studies with Video-CCAM-4B.

\begin{figure}[!t]
  \centering
  \begin{minipage}{0.49\linewidth}
    \centering
    \includegraphics[width=\linewidth]{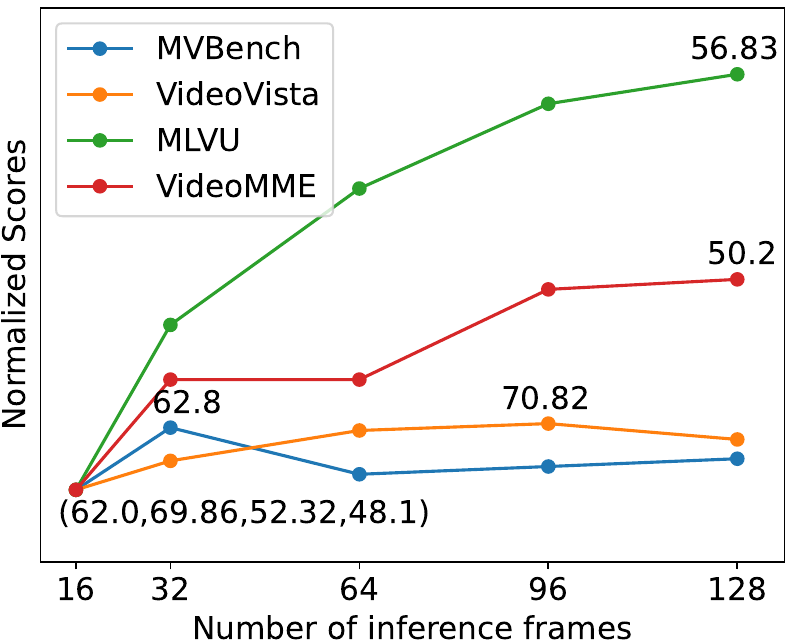}
    \subcaption{s/s$_\text{min}$-1.}
    \label{fig:6}
  \end{minipage}
  \begin{minipage}{0.49\linewidth}
    \centering
    \includegraphics[width=\linewidth]{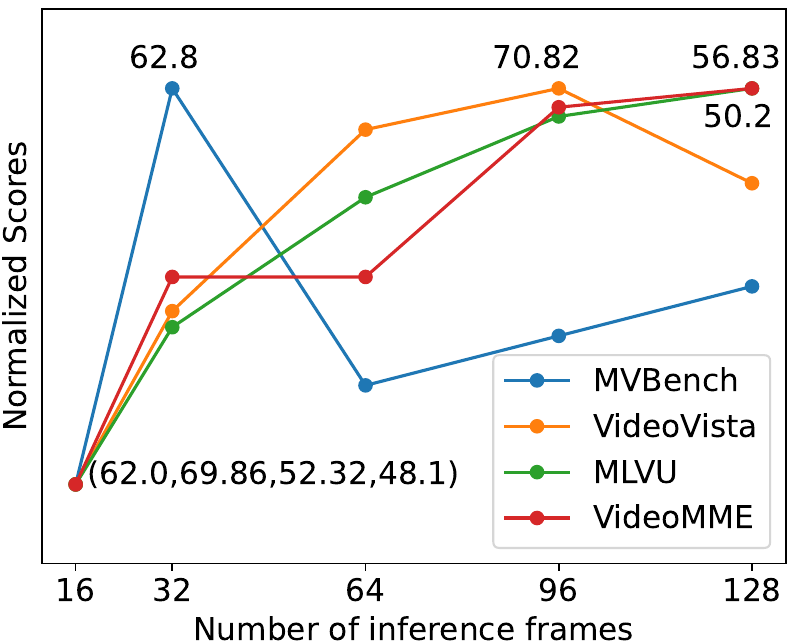}
    \subcaption{(s-s$_\text{min}$)/(s$_\text{max}$-s$_\text{min}$).}
    \label{fig:7}
  \end{minipage}
  \caption{The influence of the number of inference frames. Scores are normalized using different functions [(a),(b)] for better readability. Min/max scores are denoted in the figure.}
  \label{fig:5}
\end{figure}

\textbf{Number of Inference Frames.} We validate the temporal consistency of Video-CCAM by varying the number of inference frames in \Cref{fig:5}. In MVBench~\cite{li2024mvbench}, mostly composed of short videos, its influence is small. However, the number of inference frames plays a vital role in other benchmarks consisting of many long videos, where the score significantly increases from 16 to 96 frames and plateaus around 96 and 128 frames. Besides, no sudden improvement or degradation is observed for all benchmarks.

\textbf{CCAM.} We replace the CCAM in Video-CCAM-4B with full masks to demonstrate its necessity. We also conduct ablation studies on temporal position embeddings (TPE) as some MLLMs~\cite{li2024mvbench} use them for temporal understanding. As shown in \Cref{tab:3}, CCAM outperforms full masks by a large margin, while temporal position embeddings have a negligible impact.

\textbf{Number of Queries.} We conduct experiments by changing the number of queries to 512, 1,024, and 2,048, where Video-CCAM-4B achieves the highest score with 1,024 queries in \Cref{tab:3}. Additionally, the training duration for 2,048 queries increases by around 50\% compared to that of 1,024 queries. As a result, we settle on 1,024 learnable queries to balance performance and efficiency.

\begin{table}[htbp]
  \centering
  \caption{Ablation studies on Video-CCAM-4B.}
  \label{tab:3}
  \begin{tabular}{ccc}
    \hline
    Temporal & \#Queries & MVBench (\%) \\
    \hline
    CCAM & 1,024 & \textbf{62.80} \\
    \hline
    CCAM+TPE & 1,024 & 61.93 \\
    Full & 1,024 & 59.08 \\
    Full+TPE & 1,024 & 59.13 \\
    \hline
    CCAM & 512 & 60.78 \\
    CCAM & 2,048 & 62.68 \\
    \hline
  \end{tabular}
\end{table}

\section{Conclusion}

In this work, we introduce Video-CCAM, a novel Video-MLLM specifically designed to tackle video-language understanding tasks for both short and long videos. We integrate the causal cross-attention mask within the cross-attention layer and develop the CCAM projector to handle a large number of visual tokens and effectively model temporal dynamics. To validate its effectiveness, we conduct experiments with LLMs of different sizes on a diverse range of tasks involving both short and long videos, where Video-CCAM models consistently achieve top ranks. Our theoretical analysis and empirical studies on CCAM elucidate the factors contributing to the model's exceptional performance. Through this work, we aim to simplify the complexities of Video-MLLM development and encourage continued innovation in video-language understanding.

\bibliographystyle{plain}
\bibliography{../AuthorKit25/AnonymousSubmission/LaTeX/aaai25}



\end{document}